\documentclass[conference]{IEEEtran}
\IEEEoverridecommandlockouts
\usepackage{caption}
\usepackage{colortbl}
\usepackage{cite}
\usepackage{amsmath,amssymb,amsfonts}
\usepackage{textcomp}
\usepackage{xcolor}
\usepackage{algorithm}
\usepackage{subcaption}
\usepackage[noend]{algorithmic}
\usepackage{array}
\usepackage{multirow}
\usepackage{threeparttable}
\usepackage{booktabs}
\usepackage{stfloats}
\usepackage{url}
\usepackage{verbatim}
\usepackage{graphicx}
\usepackage{color}
\usepackage[justification=centering]{caption}

\usepackage{lettrine}
\usepackage{subcaption}
\usepackage{enumitem}

\def\BibTeX{{\rm B\kern-.05em{\sc i\kern-.025em b}\kern-.08em
    T\kern-.1667em\lower.7ex\hbox{E}\kern-.125emX}}
\begin{document}

%\title{Modality-Denoising Transformer for Enhanced Audio-Visual Zero-Shot Learning}
%\title{Modality Equilibrium Transformer for Enhanced Audio-Visual Zero-Shot Learning}
%\title{Reliance-Equilibrious Attention Network for Generalized Audio-Visual Zero-Shot Learning}
%\title{Discrepancy-Balanced Attention Network for Enhanced Audio-Visual Zero-Shot Learning}

\title{Discrepancy-Aware Attention Network for Enhanced Audio-Visual Zero-Shot Learning}
% \thanks{*contribute equally to the article\\Corresponding author: Wenrui Li}
% \author{Anonymous ICME Submission}
% arxiv版本删除了参考文献
% \author{
%     %Authors
%     % All authors must be in the same font size and format.
%     RunLin Yu, Yipu Gong, Wenrui Li, Aiwen Sun and MengRen Zheng\\

%     %Afiliations
%     Central South University, Harbin Institute of Technology, Chong Qing University\\ 
%     8208221012@csu.edu.cn;8208221221@csu.edu.cn;liwr618@163.com;\\2646416989@qq.com; 20223714@stu.cqu.edu.cn
% }
\author{\IEEEauthorblockN{RunLin Yu$^{1,\dag}$, Yipu Gong$^{1,\dag}$, Wenrui Li$^{2,*}$, Aiwen Sun$^{1}$ and Mengren Zheng$^{3}$}
\IEEEauthorblockA{$^{1}$\textit{Central South University}, $^{2}$\textit{Harbin Institute of Technology}, $^{3}$\textit{Chong Qing University} \\
\{8208221012,8208221221\}@csu.edu.cn;liwr618@163.com;2646416989@qq.com; 20223714@stu.cqu.edu.cn
}
\thanks{$^{\dag}$ Equally contributions *Corresponding author.}
}

\maketitle

\begin{abstract}
Audio-visual Zero-Shot Learning (ZSL) has attracted significant attention for its ability to identify unseen classes and perform well in video classification tasks. However, modal imbalance in (G)ZSL leads to over-reliance on the optimal modality, reducing discriminative capabilities for unseen classes. Some studies have attempted to address this issue by modifying parameter gradients, but two challenges still remain: (a) Quality discrepancies, where modalities offer differing quantities and qualities of information for the same concept. (b) Content discrepancies, where sample contributions within a modality vary significantly. To address these challenges, we propose a Discrepancy-Aware Attention Network (DAAN) for Enhanced Audio-Visual ZSL. Our approach introduces a Quality-Discrepancy Mitigation Attention (QDMA) unit to minimize redundant information in the high-quality modality and a Contrastive Sample-level Gradient Modulation (CSGM) block to adjust gradient magnitudes and balance content discrepancies. We quantify modality contributions by integrating optimization and convergence rate for more precise gradient modulation in CSGM. Experiments demonstrates DAAN achieves state-of-the-art performance on benchmark datasets, with ablation studies validating the effectiveness of individual modules.
\end{abstract}

%#######################################################################
%问题的background，是音视频联合的零样本学习。

%我们的issue，是零样本学习中对于未见类的区分能力不足。

%我们面对的problem， 是零样本学习中存在的模态不平衡，这个problem限制次优模态的贡献程度，抑制对于未见类的区分能力。

%解决模态不平衡problem中，有两个challenge，一个是音视频本身的性质不同(Quality-Discrepancy)，所含的部分冗余信息可能使得该模态占主导地位类型不同，导致模型可能过分依赖于某一个模态；二是音视频的样本质量不同(Content-Discrepancy)，同一个模态下，针对某个类别的不同样本所含有的信息，模型对其偏好也不同(不平衡)。

%为了解决challenges above, 我们提出了XXXXXXX架构，其中XXX和XXXX对应着上述的challenge。
%######################################################################
\section{Introduction}
\label{Introduction}
%Recently, Zero-Shot Learning (ZSL) in audio-visual modalities has garnered widespread attention, as it leverages the joint learning of audio and video modality features to achieve accurate classification. Many works, such as \cite{}, have demonstrated significant performance improvements over single-modality ZSL\cite{}. 
%However, many research efforts in generalized ZSL face the critical issue of modality imbalance, i.e. the disparities in the contributions that individual modalities make to the ultimate decision-making process. Due to the greedy nature of deep neural networks\cite{}, multi-modal models tend to rely heavily on high-quality modalities that contain sufficient target-related information, while underfitting the other modalities. These imbalances not only diminish the effectiveness of sub-dominant modalities but also lead to inconsistencies in convergence rates, resulting in missing modality information and suboptimal model performance\cite{}. Several studies\cite{} attempt to balance the contribution of different modalities by modulating the magnitude of gradients to slow down fast-learning modalities or accelerate slow-learning modalities, but they still fail to address some critical challenges in modality imbalance.
% Peng et al. (2022) and Xu et al. (2023).
% cite "Provable Dynamic Fusion for Low-Quality Multimodal Data" for Differential
% schonfeld2019generalized
Zero-Shot Learning (ZSL)\cite{Wang} with multi-modal audio-visual data has garnered widespread attention due to its robustness and strong generalization capabilities in capturing rich cross-modal information for complex scenarios. Its ability to handle unseen classes has seen significant improvements in the fields of video classification \cite{MDFT,10608071}. Recent studies have proposed various methods to enhance the recognition of unseen classes in audio-visual zero-shot learning, addressing a critical challenge in the field.
% work review
AVCA \cite{AVCA} enhances unseen class recognition accuracy by employing cross-attention to fuse audio-visual features, compensating for information deficiencies in each modality. TCaF \cite{tcaf} preprocesses temporal information, emphasizing its importance in the interaction between audio-visual modalities. AVMST \cite{AVMST} combines a spiking neural network for robust temporal feature extraction with a transformer reasoning module for deep feature fusion, achieving significant improvements in audio-visual zero-shot learning performance on unseen classes.
% \work review

\begin{figure}[t]
    \centering
    \includegraphics[width=1.0\columnwidth]{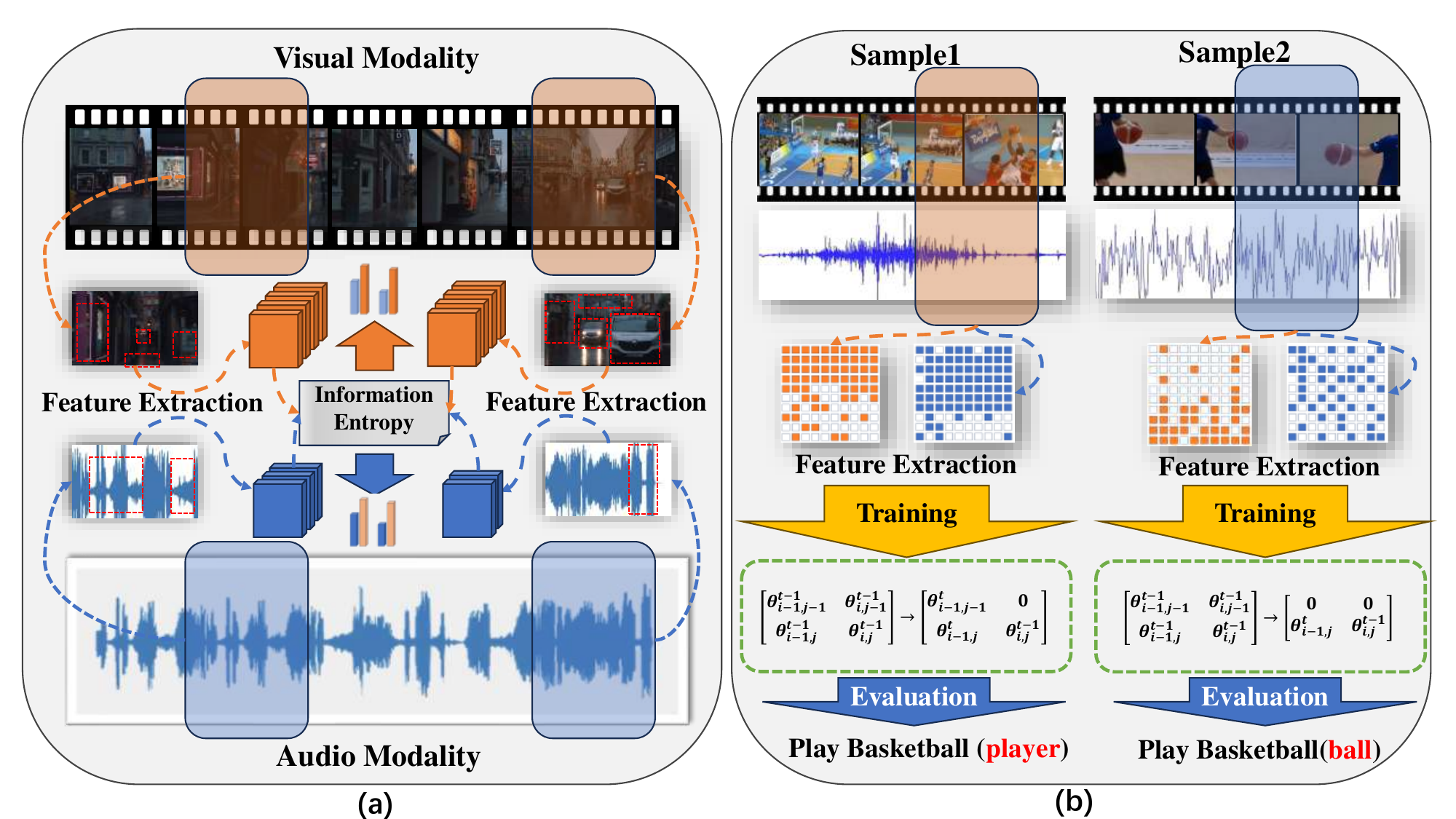}
    \captionsetup{justification=justified, singlelinecheck=false}
    \caption{\small{Quality \& Content Discrepancies in Audio-Visual Dataset: (a) Quality discrepancies exist between audio and visual modalities. The visual one holds more target-related data, causing the model's prediction to depend more on it. (b) Content discrepancies occur in samples. For the same category like playing basketball, distinct audio-visual samples result in diverse recognition outcomes because of different information biases (Sample 1 emphasizes the player, while Sample 2 emphasizes the ball).}}
    \label{fig1}
\end{figure}

However, these studies ignore the problem of modality imbalance in multi-modal learning, which refers to the unequal contributions of individual modalities to the final decision-making process\cite{MultiModimbalance,what,ismail2020improving,sun2021learning}. Due to the greedy nature of deep neural networks\cite{greedy}, multi-modal models often over-rely on high-quality modalities that contain sufficient target-related information, while underfitting the other modalities. These imbalances reduce the effectiveness of sub-dominant modalities and cause inconsistencies in convergence rates, leading to missing modality information and suboptimal model performance \cite{greedy}. Various methods have been proposed to address the imbalance in multi-model joint learning. OGM-GE\cite{OGM} method is a dynamic gradient modulation strategy that adjusts the gradient of unimodal through the discrepancy in the accuracy of their all samples. However, the dominant modality not only restricts the learning rates of other modalities but also disrupts their update directions. PMR\cite{PMR} introduces a prototype-based entropy regularization term during the early training stage to prevent premature convergence. Furthermore, to evaluate the fine-grained contribution of each modality, Wei et al. \cite{samplelevel} propose a sample-level modality valuation metric to assess and enhance fine-grained cooperation among modalities, which is particularly useful for multi-modal systems involving numerous modalities. Although previous methods have effectively alleviated the impact of modal imbalance, there are still two critical challenges in (G)ZSL:
(a) \textbf{\emph{Quality Discrepancies}} in audio-visual modalities, where different modalities contain varying amounts and qualities of information when describing the same concept \cite{multimodalsurvey}. (b) \textbf{\emph{Content Discrepancies}} in sample-level modalities, where the contributions of different samples within the same modality exhibit significant differences \cite{samplelevel}.

Specifically, quality discrepancies indicate that visual data may contain more information related to the target event or object compared to audio data. This difference may cause deep neural networks to rely on the high-quality modality that provides sufficient target-related information while neglecting other modalities. As shown in Fig. \ref{fig1}(a), modalities contributing more critical features have greater opportunities for optimization, leaving other modalities under-optimized. Meanwhile, content discrepancies describes the contributions of different samples of the same modality to the final prediction can fluctuate significantly. In Fig. \ref{fig1}(b), the same category (basketball) still produces varying outcomes due to significant information discrepancies. These discrepancies are challenging for existing models to capture, as they typically focus on global modality discrepancies and fail to address fine-grained modality contribution discrepancies within each sample \cite{multimodalsurvey}. Content discrepancies are critical for understanding and enhancing the performance of multimodal learning models, as they directly influence how the model utilizes information from various modalities for predictions.

% more specific on methods.
To address the aforementioned challenges, we propose a Discrepancy-Aware Attention Network (DAAN) for Enhanced Audio-Visual Zero-Shot Learning. Specifically, we propose a Quality Discrepancy Mitigation Attention (QDMA) unit to reduces the redundant information contained in the high-quality modality. In order to form a sparse attention, QDMA calculates two independent softmax attention scores of the same modality and subtracts the two scores with weights to obtain the softmax scores of the sparse attention. Meanwhile, QDMA utilizes the Temporal Convolutional Network (TCN)\cite{TCN} to extract the temporal embeddings of audio-visual features so as to enhance the temporal information.

To eliminate the impact of existing content discrepancies among samples, we design a Contrastive Sample-Level Gradient Modulation (CSGM) block. The block adjusts the magnitude of the model gradient based on the contribution rate. Specifically, CSGM operates on each pair of positive and negative samples, rather than applying uniform modulation across multiple epochs based on the mean accuracy of modality features. This allows for capturing and balancing the content discrepancy at the sample level. 
To ensure precise gradient modulation, we introduce a novel method to quantify modality contributions, defining the contribution rate as a combination of the optimization rate and the convergence rate. Inspired by \cite{greedy}, we utilize the ratio of the squared 2-norm of the module gradient to that of the parameter to represent the optimization rate. The convergence rate is calculated by measuring the degree of aggregation between positive and negative samples across different modalities and projecting it within a defined value range.

The main contributions of this paper are as follows:
\begin{itemize}
    % \item We propose an DAAN architecture to simultaneously balance the learning of audio-visual modalities. 
    %Moreover, our architecture is the first to consider both Quality and Content Discrepancies at the same time.
    % \item We propose a QDMA unit, which reduces the redundant information contained in the high-quality modality and enhances ZSL performance.
    \item We propose a QDMA unit, which mitigates quality discrepancies by reducing redundancy in high-quality modalities through a sparse attention mechanism and enhances temporal information using TCN.
    \item A CSGM block is proposed to dynamically modulate the gradients magnitude of parameters and balance content discrepancy in a sample level.
    \item We develop an novel approach to quantify modality contributions by integrating their optimization and convergence for more accurate gradient modulation in CSGM.
    \item  The DAAN achieves state-of-the-art performance across VGGSound, UCF101, and ActivityNet datasets. Additionally, ablation studies prove the effectiveness of different modules within our architecture.
\end{itemize}
 
\section{Method}
\label{Method}

\begin{figure*}[!h]
    \centering
    \includegraphics[width=2.0\columnwidth]{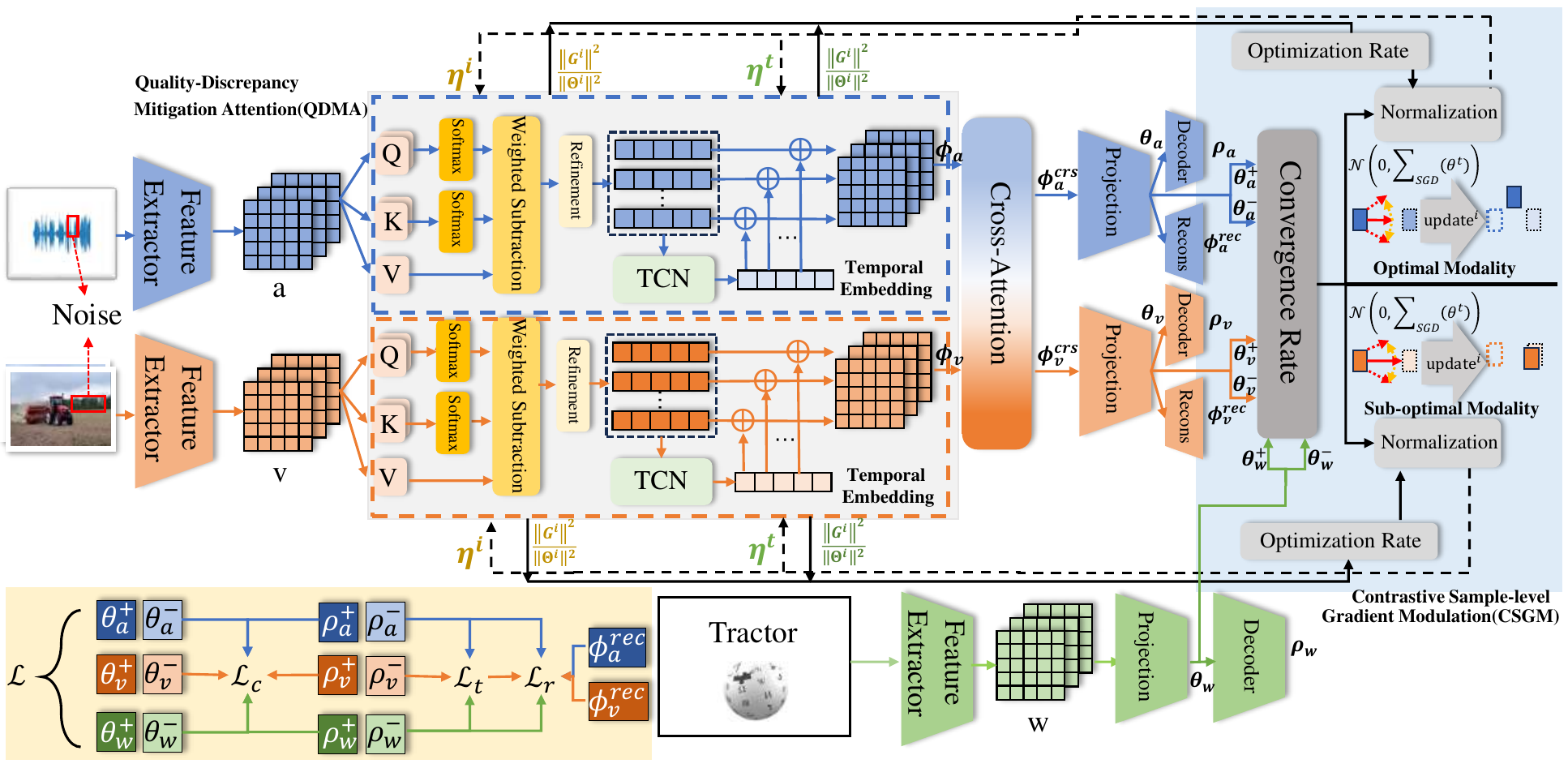}
    \captionsetup{justification=justified, singlelinecheck=false}
    \caption{\small{The DAAN architecture incorporates audio, visual, and textual features as inputs, simultaneously extracting semantic information. The QDMA unit removes redundant information from high-quality modalities to address quality discrepancies. Additionally, it extracts temporal embeddings of audio-visual features to enhance temporal information. The cross-attention layer facilitates information interaction between audio-visual features. The CSGM block dynamically adjusts the parameter gradients of various modules within the QDMA unit at the sample level, effectively eliminating content discrepancies among samples. The loss functions are illustrated in the lower-left area.}}
    \label{fig2}
\end{figure*}

The extracted audio and visual features are denoted as \(\boldsymbol{a}_{i}\) and \(\boldsymbol{v}_{i}\) respectively, and the text feature of the corresponding labelled classes is denoted as \(\boldsymbol{w}_{i}\). In (G)ZSL, we define the seen class as \(\mathcal{S}=\{\boldsymbol{v}_{i}^{s}, \boldsymbol{a}_{i}^{s}, \boldsymbol{l}_{i}\}\) and the unseen class as \(\mathcal{U}=\{\boldsymbol{v}_{i}^{u}, \boldsymbol{a}_{i}^{u}, \boldsymbol{l}_{i}\}\), where \(\boldsymbol{l}_{i}\) represents the corresponding \(i_{th}\) class label. The DAAN architecture is proposed to learn the projection function \(\mathcal{F}: (\boldsymbol{v}_{i}^{s}, \boldsymbol{a}_{i}^{s})\rightarrow \boldsymbol{w}_{i}^{s}\), which can also be applied to unseen class \(\mathcal{F}: (\boldsymbol{v}_{i}^{u}, \boldsymbol{a}_{i}^{u})\rightarrow \boldsymbol{w}_{i}^{u}\). The architecture of DAAN is shown in Fig. (\ref{fig2}).

\subsection{\textbf{Quality-Discrepancy Mitigation Attention (QDMA)}} To minimize the impact of quality discrepancies between the audio and visual modalities, features \(\boldsymbol{a}_{i}\) and \(\boldsymbol{v}_{i}\) acquired from pre-trained feature extractor would be simultaneously input into the QDMA unit. Specifically, \(\boldsymbol{a}_{i}\) and \(\boldsymbol{v}_{i}\) are respectively split into two feature tensors, denoted as \((\boldsymbol{a}_{i}^{1}, \boldsymbol{a}_{i}^{2})\in\mathbb{R}^{\frac{input}{2}}\) and \((\boldsymbol{v}_{i}^{1}, \boldsymbol{v}_{i}^{2})\in\mathbb{R}^{\frac{input}{2}}\). Afterwards, \((\boldsymbol{a}_{i}^{1}, \boldsymbol{a}_{i}^{2})\), \((\boldsymbol{v}_{i}^{1}, \boldsymbol{v}_{i}^{2})\) and \(\boldsymbol{w}_{i}\) are multiplied by the corresponding matrices \(\boldsymbol{W}_{i}^{Q}\), \(\boldsymbol{W}_{i}^{K}\) and \(\boldsymbol{W}_{i}^{V}\) respectively, and the obtained logits \((\boldsymbol{Q}_{1}^{i}, \boldsymbol{Q}_{2}^{i}), (\boldsymbol{K}_{1}^{i}, \boldsymbol{K}_{2}^{i})\) and \(\boldsymbol{V}^{i}\) would undergo the softmax \(S(\cdot)\). Inspired by the common-mode noise cancellation in analog circuits \cite{diff}, we perform weighted subtraction on the different softmax values from the same feature to eliminate the redundant information contained in the high-quality modal features while enhancing the key information, that is: 
\begin{equation}
    \boldsymbol{o}_{i}^{1} = S(\boldsymbol{Q}_{1}^{i}{\boldsymbol{K}_{1}^{i}}^{T}) - \beta\cdot S(\boldsymbol{Q}_{2}^{i}{\boldsymbol{K}_{2}^{i}}^{T})\boldsymbol{V}^{i},
\end{equation}
where \(\beta\) is a hyper-parameter that provides flexibility to the QDMA block. Subsequently, the audio-visual features pass through the refinement layer \(f^{r}: \mathbb{R}^{input}\rightarrow\mathbb{R}^{hidden}\), which comprises of GroupNorm\cite{groupnorm} and a linear layer followed by BatchNorm, ReLu and Dropout. It is represented as:
\begin{equation}
\begin{aligned}
    \boldsymbol{o}_{i}^{2} &= GrpNorm(\boldsymbol{o}_{i}^{1}),\\
    \boldsymbol{o}_{i}^{3} &= (1 - \beta)\cdot Linear(\boldsymbol{o}_{i}^{2}),
\end{aligned}
\end{equation}
where \(\boldsymbol{o}_{i}^{2}\) and \(\boldsymbol{o}_{i}^{3}\in\mathbb{R}^{hidden}\) represent the output features corresponding to GroupNorm and Linear respectively.

To further enhance the temporal information of audio and visual features, we have designed a temporal bypass based on the TCN. Unlike conventional sequential networks, the TCN mainly consists of the Causal Convolutional Layer \(\mathcal{C}\), \(\boldsymbol{n}\) layers of Dilated Convolution \(\mathcal{D}\) and Residual Connection \(\mathcal{R}\). We define the corresponding weight parameters as \(\boldsymbol{w}^{\mathcal{C}}\), \(\boldsymbol{w}^{\mathcal{D}}\) and \(\boldsymbol{w}^{\mathcal{R}}\) The outputs of TCN can be written as follows:
\begin{equation}
\begin{aligned}
    \boldsymbol{y}_{i}^{t+1} = \mathcal{R}(\mathcal{D}(\mathcal{C}(\boldsymbol{o}_{i}^{3}))),
\end{aligned}
\end{equation}
where \(\boldsymbol{o}_{i}^{3}\) is reshaped as \(\mathbb{R}^{hidden}\rightarrow\mathbb{R}^{x^{hidden}\times 1\times y^{hidden}}\).

In the aforementioned process, \(\mathcal{C}\) uses one-dimensional convolution to process the feature sequence, denoted as \(\boldsymbol{y}_{j,i}^{t-1} = \sum_{n=0}^{y^{hid} - 1}(\boldsymbol{x}_{j-n,i}^{t-1}\cdot \boldsymbol{w}_{n}^{\mathcal{C}})\), where \(\boldsymbol{x}_{j-n,i}^{t-1}\subseteq\boldsymbol{o}_{i}^{t+1}, \boldsymbol{x}_{j-n,i}^{t-1}\in\mathbb{R}^{1*y^{hid}}\). It ensures that the network does not utilize the information of future time steps when predicting the output at the current time step. Then, \(\mathcal{D}\) with \(\boldsymbol{n}\) layers increases the receptive field by inserting gaps \(\boldsymbol{k}\) (i.e., dilation) among the elements of the convolution kernel, which is \(\boldsymbol{y}_{j,i}^{t} = \sum_{n=0}^{y^{hid} - 1}(\boldsymbol{x}_{j-n\cdot k,i}^{t-1}\cdot \boldsymbol{w}_{n}^{\mathcal{D}})\). Lastly, \(\mathcal{R}\) adds the input directly to the output of the convolution layer to alleviate the problem of vanishing gradients and enhance the learning ability and stability of the network. It is defined as \(\boldsymbol{y}_{j,i}^{t+1} = \mathcal{D}(\mathcal{C}(\boldsymbol{x}_{j,i}^{t+1}))+\boldsymbol{x}_{j,i}^{t+1}\). We take the temporal feature at the last time step of the \(\boldsymbol{y}_{i}^{t+1}\) as the temporal embedding and add it to the audio and visual feature tensors, described as:
\begin{equation}
    \boldsymbol{\phi}_{i} = \sum_{l=0}^{x^{hid}}(\boldsymbol{o}_{l,i}^{t+1} + \boldsymbol{y}_{x^{hid},i}^{t+1}), \boldsymbol{\phi}_{i}\in\{\boldsymbol{\phi}_{a,i}, \boldsymbol{\phi}_{v,i}\}.
\end{equation}

%improve the recognition ability for unseen classes as well as the generalization performance of the model
To facilitate information sharing between the audio and visual modalities, we utilize a cross-attention block to process the features \(\boldsymbol{\phi}_{a,i}\) and \(\boldsymbol{\phi}_{v,i}\) after eliminating quality discrepancies. Cross-attention primarily comprises a multi-head self-attention layer and a fully-connected feed-forward block. In the multi-head self-attention layer, query, key, and value tensors are derived by cross-utilizing audio and visual features, enabling effective information sharing and fusion across modalities. The fully-connected feed-forward block further processes these integrated features. Finally, the integrated features are merged with the original features via residual connections and layer normalization. The resulting output features are represented as \(\boldsymbol{\phi}_{a,i}^{crs}\in\mathbb{R}^{hidden}\) and \(\boldsymbol{\phi}_{v,i}^{crs}\in\mathbb{R}^{hidden}\).

\subsection{\textbf{Contrastive Sample-level Gradient Modulation (CSGM)}}
In order to align the output features \(\boldsymbol{\phi}_{a,i}^{crs}\) and \(\boldsymbol{\phi}_{v,i}^{crs}\) with the text feature \(\boldsymbol{w}_{i}\), we utilize a projection layer to project all three into the same tensor space, respectively denoted as \(\boldsymbol{\theta}_{a}^{i}\), \(\boldsymbol{\theta}_{v}^{i}\) and \(\boldsymbol{\theta}_{w}^{i} \in \mathbb{R}^{output}\). The projection layer consists of a linear layer followed by BatchNorm, ReLU, and Dropout. 

Subsequently, to address content discrepancies in audio-visual samples, we introduce the CSGM block, which comprehensively modulates gradient magnitudes across different modules. Specifically, gradient modulation is performed across different modules based on their respective contribution rates, represented as \(\boldsymbol\eta^{i}\). \(\boldsymbol{\eta}^{i}\) is calculated from the optimization rate of the model and the convergence rate of the corresponding modality, and is expressed as:
\begin{equation}
\begin{aligned}
    \boldsymbol{\eta}_{a,p}^{i} &= max(\boldsymbol{\mathcal{V}}_{a,c}^{i} \times \boldsymbol{\mathcal{V}}_{a,o,p}^{i}, \boldsymbol{\gamma}),\\
    \boldsymbol{\eta}_{v,p}^{i} &= max(\boldsymbol{\mathcal{V}}_{v,c}^{i} \times \boldsymbol{\mathcal{V}}_{v,o,p}^{i}, \boldsymbol{\gamma}),
\end{aligned}
\end{equation}
where \(\boldsymbol{\mathcal{V}}_{c}^{i}\) and \(\boldsymbol{\mathcal{V}}_{o,p}^{i}\) (\(p\in\{1,2\}\)) respectively represent the convergence rate and optimization rate that are normalized under the regulation of weight parameter \(\boldsymbol{\mu}\) in sample \(i\). . \(\boldsymbol{\gamma}\) is a hyper-parameter that is used to prevent an excessively minute modality gradient, which brings flexibility to the model. 

Considering the information discrepancies in the features corresponding to distinct samples, we leverage the features from each group of positive and negative samples to perform contrastive learning for calculating \(\boldsymbol{\mathcal{V}}_{c}^{i}\). We define \((\boldsymbol{\theta}_{a,i}^{+},\boldsymbol{\theta}_{a,i}^{-})\), \((\boldsymbol{\theta}_{v,i}^{+},\boldsymbol{\theta}_{v,i}^{-})\) and \((\boldsymbol{\theta}_{w,i}^{+},\boldsymbol{\theta}_{w,i}^{-})\) as the sets of positive and negative embeddings for audio, visual and text in the same tensor space, respectively, and use the Euclidean distance \(d(x, y) = \lVert x - y\rVert_{2}\) to calculate the similarity between embeddings. For each modality, we calculate the convergence of the corresponding positive and negative embeddings respectively, which is expressed as \(\boldsymbol{\delta}_{m,e,n}^{i} = max(d(\boldsymbol{\theta}_{m,i}, \boldsymbol{\theta}_{e,i}) - d(\boldsymbol{\theta}_{m,i}, \boldsymbol{\theta}_{n,i}),0), m,e,n\in\{a^{+},v^{+},w^{+},a^{-},v^{-},w^{-}\}\). Therefore, the convergence rate for audio-visual modalities can be calculated as follows:
% l_taw = self.triplet_loss(self.theta_a, self.theta_w, self.theta_a_neg)
% l_ta = self.triplet_loss(self.theta_w, self.theta_a, self.theta_a_neg)
% l_at = self.triplet_loss(self.theta_a, self.theta_w, self.theta_w_neg)
% l_tvw = self.triplet_loss(self.theta_v, self.theta_w, self.theta_v_neg)
% l_tv = self.triplet_loss(self.theta_w, self.theta_v, self.theta_v_neg)
% l_vt = self.triplet_loss(self.theta_v, self.theta_w, self.theta_w_neg)
\begin{equation}
\begin{aligned}
    \boldsymbol{\mathcal{V}}_{a,c}^{i} &= \boldsymbol{\delta}_{a^{+},w^{+},a^{-}}^{i} \cdot \boldsymbol{\delta}_{w^{+},a^{-},a^{-}}^{i} \cdot \boldsymbol{\delta}_{a^{+},w^{-},w^{-}}^{i}, \\
    \boldsymbol{\mathcal{V}}_{v,c}^{i} &= \boldsymbol{\delta}_{v^{+},w^{+},v^{-}}^{i} \cdot \boldsymbol{\delta}_{w^{+},v^{-},v^{-}}^{i} \cdot \boldsymbol{\delta}_{v^{+},w^{-},w^{-}}^{i}.
\end{aligned}
\end{equation}

In the abovementioned formulas, \(\boldsymbol{\delta}_{m,e,n}^{i}\) are all normalized to prevent the explosion of the modulation amplitude. To alleviate the biases of the QDMA unit caused by the content discrepancies of samples, we also calculate the corresponding \(\boldsymbol{\mathcal{V}}_{o}^{i}\) for the parameters of the same module under different modalities. Inspired by \cite{greedy}, we define \(\boldsymbol{\mathcal{G}}_{p}^{i} = \frac{\partial\boldsymbol{\mathcal{L}}}{\partial\boldsymbol{\Theta}_{p}^{i-1}}\), where \(\boldsymbol{\Theta}_{p}^{i-1}\) represents the learnable parameters of part \(p\) in QDMA. Hence, \(\boldsymbol{\mathcal{V}_{o}^{i}}\) is calculated as follows:
\begin{equation}
\begin{aligned}
    \boldsymbol{\mathcal{V}}_{a,o,p}^{i} &= \frac{\lVert\boldsymbol{\mathcal{G}}_{p}^{i}\rVert_{2}^{2}}{\lVert\boldsymbol{\Theta}_{p}^{i-1}\rVert_{2}^{2}},\\
    \boldsymbol{\mathcal{V}}_{v,o,p}^{i} &= \frac{\lVert\boldsymbol{\mathcal{G}}_{p}^{i}\rVert_{2}^{2}}{\lVert\boldsymbol{\Theta}_{p}^{i-1}\rVert_{2}^{2}},
\end{aligned}
\end{equation}
where \(\lVert\cdot\rVert_{2}^{2}\) demonstrates the square of the 2-norm. In conclusion, CSGM can be expressed as:
\begin{equation}
\begin{aligned}
    \boldsymbol{\Theta}_{a,p}^{i} &=  \boldsymbol{\Theta}_{a,p}^{i-1} - \boldsymbol{\eta}_{a,p}^{i}\cdot\boldsymbol{\mathcal{G}}_{a,p}^{i} + \boldsymbol{\epsilon}^{i},\\
      \boldsymbol{\Theta}_{v,p}^{i} &=  \boldsymbol{\Theta}_{v,p}^{i-1} - \boldsymbol{\eta}_{v,p}^{i}\cdot\boldsymbol{\mathcal{G}}_{v,p}^{i} + \boldsymbol{\epsilon}^{i}.
\end{aligned}
\end{equation}

It is worth noting that \(\boldsymbol{\epsilon}^{i} \sim \mathcal{N}(0, \sum_{SGD}\Theta)\) is the Gaussian noise term added during modulation to enhance the generalization ability of our modal. 

\subsection{\textbf{Training Strategy}}\label{Training Strategy}
To ensure decoding text label embeddings from audio-visual embeddings for feature alignment and encourage alignment of different modality embeddings while retaining input modality information, we employ reconstructors for audio-visual features and decoders for all three features, respectively denoted as \((\boldsymbol{\rho}_{a}^{i}, \boldsymbol{\phi}_{a,i}^{rec})\), \((\boldsymbol{\rho}_{v}^{i}, \boldsymbol{\phi}_{v,i}^{rec})\) and \(\boldsymbol{\rho}_{w}^{i}\). During the training, we update the modal's parameters using the loss function \(\boldsymbol{\mathcal{L}}\), which comprises of triplet loss \(\boldsymbol{\mathcal{L}}_{t}\), composite triplet and reconstruction loss \(\boldsymbol{\mathcal{L}}_{c}\) and regularization loss \(\boldsymbol{\mathcal{L}}_{r}\). 

\noindent\textbf{Triplet Loss} The triplet loss enables the features of different classes to be separated as much as possible in the feature space, and improves the model's ability to distinguish between different classes, especially the classification accuracy for unseen classes. It can be written as:
\begin{equation}
\begin{aligned}
    \boldsymbol{\mathcal{L}}_{t}&= t(\boldsymbol{\theta}_{a,i}^{+},\boldsymbol{\theta}_{w,i}^{+},\boldsymbol{\theta}_{a,i}^{-}) + t(\boldsymbol{\theta}_{v,i}^{+},\boldsymbol{\theta}_{w,i}^{+},\boldsymbol{\theta}_{v,i}^{-})\\ &+t(\boldsymbol{\theta}_{w,i}^{+},\boldsymbol{\theta}_{a,i}^{+},\boldsymbol{\theta}_{w,i}^{-}) +t(\boldsymbol{\theta}_{w,i}^{+},\boldsymbol{\theta}_{v,i}^{+},\boldsymbol{\theta}_{w,i}^{-}),
\end{aligned}
\end{equation}
where \(t(\cdot)\) represents the triplet loss function.

\noindent\textbf{Composite Loss} The composite loss \(\boldsymbol{\mathcal{L}}_{c}\), comprising the reconstruction loss \(\boldsymbol{l}_{rec}\) and triplet loss \(\boldsymbol{l}_{ct}\), is employed in our modal. \(\boldsymbol{l}_{rec}\) ensures decodability by minimizing the mean squared error (MSE) between decoded features and text embeddings, bridging audio-visual and textual information. \(\boldsymbol{l}_{ct}\) and \(\boldsymbol{l}_{w}\) align features from different modalities by leveraging sample distances, thereby enhancing feature representations, facilitating multimodal fusion, and improving classification for unseen classes. The specific definition is as below:
\begin{equation}
\begin{aligned}
    \boldsymbol{l}_{rec} &= d(\boldsymbol{\rho}_{a}^{i}, \boldsymbol{w}_{i}) + d(\boldsymbol{\rho}_{v}^{i}, \boldsymbol{w}_{i}) + d(\boldsymbol{\rho}_{w}^{i}, \boldsymbol{w}_{i}),\\
    \boldsymbol{l}_{ct} &= t(\boldsymbol{\rho}_{w,i}^{+},\boldsymbol{\rho}_{a,i}^{+}, \boldsymbol{\rho}_{a,i}^{-}) + t(\boldsymbol{\rho}_{w,i}^{+},\boldsymbol{\rho}_{v,i}^{+}, \boldsymbol{\rho}_{v,i}^{-}),\\
    \boldsymbol{l}_{w} &= t(\boldsymbol{\theta}_{w,i}^{+},\boldsymbol{\theta}_{a,i}^{+}, \boldsymbol{\theta}_{a,i}^{-}) + t(\boldsymbol{\theta}_{w,i}^{+},\boldsymbol{\theta}_{v,i}^{+}, \boldsymbol{\theta}_{v,i}^{-})\\
    &+t(\boldsymbol{\theta}_{a,i}^{+},\boldsymbol{\theta}_{w,i}^{+}, \boldsymbol{\theta}_{w,i}^{-}) +t(\boldsymbol{\theta}_{v,i}^{+},\boldsymbol{\theta}_{w,i}^{+}, \boldsymbol{\theta}_{w,i}^{-}),\\
\end{aligned}
\end{equation}
and the composite loss is written as \(\boldsymbol{\mathcal{L}_{c}=\boldsymbol{l}_{rec}+\boldsymbol{l}_{ct}}+\boldsymbol{l}_{w}\).

\noindent\textbf{Regularization Loss} Lastly, we utilize a regularization loss to promote the alignment of audio and visual embeddings with text embeddings, while preserving the distinctive information of respective input modalities. This enables adaptation to varying data distributions, particularly for unseen class data. It is defined as:
\begin{equation}
\begin{aligned}
    \boldsymbol{l}_{r} &= d(\boldsymbol{\phi}_{a,i}^{rec}, \boldsymbol{\phi}_{a,i}) + d(\boldsymbol{\phi}_{v,i}^{rec}, \boldsymbol{\phi}_{v,i})\\
    &+ d(\boldsymbol{\phi}_{a}^{i}, \boldsymbol{\phi}_{w}^{i}) + d(\boldsymbol{\phi}_{v}^{i}, \boldsymbol{\phi}_{w}^{i}).
\end{aligned}
\end{equation}
In all, the total loss is defined as:
\begin{equation}
\boldsymbol{\mathcal{L}=\boldsymbol{\mathcal{L}}_{t}+\boldsymbol{\mathcal{L}}_{c}}+\boldsymbol{\mathcal{L}}_{r}.
\end{equation}

\section{Experiments and Results}
\label{Experiments and Results}
\subsection{\textbf{Experimental Settings}}\label{Experimental Settings}
\begin{table*}[!h]
	\centering
  \renewcommand\arraystretch{1.5}
	\begin{threeparttable}  
		\caption{The performance of our DAAN and state-of-the-art baselines for audio-visual (G)ZSL on three benchmark datasets.}
		\label{tab1}
		\setlength{\tabcolsep}{6pt}{
\begin{tabular}{cccccclcccclcccc}
\hline \hline
\multirow{2}{*}{Type} &
  \multirow{2}{*}{Model} &
  \multicolumn{4}{c}{VGGSound-GZSL} &
   &
  \multicolumn{4}{c}{UCF-GZSL} &
   &
  \multicolumn{4}{c}{ActivityNet-GZSL} \\ \cline{3-6} \cline{8-11} \cline{13-16} 
 &
   &
  S &
  U &
  \textit{HM $\uparrow$} &
  \textit{ZSL $\uparrow$} &
   &
  S &
  U &
  \textit{HM $\uparrow$} &
  \textit{ZSL $\uparrow$} &
   &
  S &
  U &
  \textit{HM $\uparrow$} &
  \textit{ZSL $\uparrow$} \\ \hline
\multirow{5}{*}{\begin{tabular}[c]{@{}c@{}}Audio-visual\\ ZSL\end{tabular}} &
CJME \cite{CJME} &
  8.69 &
  4.78 &
  6.17 &
  5.16 &
   &
  26.04 &
  8.21 &
  12.48 &
  8.29 &
   &
  5.55 &
  4.75 &
  5.12 &
  5.84 \\
 &
  AVGZSLNet \cite{AVGZSLNet} &
  18.05 &
  3.48 &
  5.83 &
  5.28 &
   &
  52.52 &
  10.90 &
  18.05 &
  13.65 &
   &
  8.93 &
  5.04 &
  6.44 &
  5.40 \\
 &
  TSART \cite{TSART} &
  10.45 &
  3.43 &
  5.16 &
  4.03 &
   &
  20.96 &
  21.27 &
  21.11 &
  22.86 &
   &
  8.99  &
  7.41 &
  8.12 &
  7.65 
  \\&
  AVCA \cite{AVCA} &
  14.90 &
  4.00 &
  6.31 &
  6.00 &
   &
  51.53 &
  18.43 &
  27.15 &
  20.01 &
   &
  24.86 &
  8.02 &
  12.13 &
  9.13 \\
 &
  TCaF \cite{tcaf} &
  9.64 &
  5.91 &
  7.33 &
  6.06 &
   &
  58.60 &
  21.74 &
  31.72 &
  24.81 &
   &
  18.70 &
  7.50 &
  10.71 &
  7.91 \\ 
 &
  AVMST \cite{AVMST}&
  14.14 &
  5.28 &
  7.68 &
  6.61 &
   &
  44.08 &
  22.63 &
  29.91 &
  28.19 &
   &
  17.75 &
  9.90 &
  12.71 &
  10.37 \\& 
Hyper$^{\mathrm{alignment}}$ \cite{Hyper-multiple} &
  13.22 &
  5.01 &
  7.27 &
  6.14 &
   &
  57.28 &
  17.83 &
  27.19 &
  19.02 &
   &
  23.50 &
  8.47 &
  12.46 &
  9.83
 \\&
    MDFT \cite{MDFT} &
  16.14 &
  5.97 &
  8.72 &
  7.13 &
   &
  48.79 &
  23.11 &
  31.36 &
  \textbf{31.53} &
   &
  18.32 &
  10.55 &
  13.39 &
  \textbf{12.55} 
  \\
%     STFT \cite{STFT} &
%   19.22 &
%   6.81 &
%   \textbf{10.06}&
%   \textbf{8.24} &
%    &
%   56.47 &
%   22.89 &
% \textbf{32.58}&
%   29.72 &
%    &
%   22.34 &
%   11.73 &
%   \textbf{15.38} &
%   \textbf{12.91} \\
  \hline \rowcolor{lightgray!30}
& DAAN (ours) &
  10.04 &
  7.10 &
  8.32 &
  \textbf{7.91} &
   &
  52.38 &
  23.48 &
  \textbf{32.42} &
  31.09 &
   &
  20.91 &
  10.26 &
  \textbf{13.87} &
  11.15 \\
\hline \hline
\end{tabular}
}
\end{threeparttable}
\end{table*}

In this paper, we assess our proposed model in (G)ZSL scenarios using the VGGSound \cite{VGG}, UCF101 \cite{UCF}, and ActivityNet \cite{Activitynet} datasets. Consistent with the approach in\cite{AVCA}\cite{benchmark}, we employ mean class accuracy to measure the model's effectiveness in classification tasks. For the ZSL evaluation, we concentrate on the model's performance on test samples from unseen classes. For the GZSL evaluation, we evaluate the model on both seen (\(\mathcal{S}\)) and unseen (\(\mathcal{U}\)) classes. To balance the performance between seen and unseen classes and offer an overall evaluation of modal performance in GZSL, we deploy harmonic mean ($HM$) metrics, denoted as \({HM} = \frac{2\cdot\mathcal{US}}{\mathcal{U}+\mathcal{S}}\). 

For each video sample, we use the self-supervised SeLaVi\cite{selavi} framework pretrained on those datasets to extract audio and visual features for each second in a video. DAAN are then trained on a single NVIDIA 3090 GPU, and the dimensions are respectively set to $input = 512$, $hidden = 512$, $output = 300$. \(\boldsymbol{\mu}\) is set to $1.15/0.5/1.2$, \(\boldsymbol{\gamma}\) is set to $0.45/0.5/0.6$, \(\boldsymbol{k}\) is set to $3/9/5$, \(\boldsymbol{n}\) is set to $2/3/5$ for UCF/ActivityNet/VGGSound respectively. All models are trained $50$ epochs using Adam optimize with a pre-defined learning rate of $0.001$.

\subsection{\textbf{Comparison with State-of-the-Art}}
To demonstrate the advancement of our work, we compare the DAAN with the following state-of-the-art audio-visual (G)ZSL frameworks: CJME\cite{CJME}, AVGZSLNet\cite{AVGZSLNet}, AVCA\cite{AVCA}, TCaF\cite{tcaf}, Hyper$^{\mathrm{alignment}}$ \cite{Hyper-multiple} and MDFT\cite{MDFT}. 
 
Table \ref{tab1} shows that DAAN achieves the level of state-of-the-art performance across VGGSound, UCF101, and ActivityNet datasets on (G)ZSL performance, significantly exceeding almost all baselines.It is worth noting that although DAAN does not surpass the latest state-of-the-art MDFT in some metrics, the runtime of MDFT is $2.5$ times more than DAAN due to its use of the Spiking Neural Network, which suggests that DAAN is significantly more efficient than MDFT. Moreover, our model greatly exceeds the best performance of baseline modal. For UCF-GZSL, DAAN outperforms the best previous method AVMST with an HM of $32.42$ compared to $29.91$ and a ZSL of $31.09$ compared to $28.19$. This demonstrates that the DAAN effectively integrates the functions of QDMA and CSGM, alleviating the impact of modality imbalance and enhancing the generalization ability of the model.  

\subsection{\textbf{Ablation Study}}\label{tab2}
We conduct ablation experiments on the UCF dataset to illustrate the effectiveness individual component in our model, as shown in Table \ref{tab:ablation_ucf-gzsl}. Among those components, QDMA brings a $1.03\%$ increase in HM and a $66.12\%$ increase in ZSL, and CSGM using convergence rate as the only metric gets a $18.33\%$ increase in HM and a slightly $1.19\%$ drop in ZSL. The complete CSGM with both convergence rate and optimization rate reaches a $22.52\%$ increase in HM and a $8.03\%$ increase in ZSL. Lastly, our DAAN holds a $25.81\%$ increase in HM and a $63.37\%$ increase in ZSL compared to “Base” in total.

\begin{table}[t]
	\centering
  \renewcommand\arraystretch{1.3}
	\caption{Ablation study on UCF-GZSL.}
	\label{tab:ablation_ucf-gzsl}
	\setlength{\tabcolsep}{6pt}{
\begin{threeparttable}
\begin{tabular}{lcccc}
\hline \hline
\multirow{2}{*}{Model} &
  \multicolumn{4}{c}{UCF-GZSL} \\
  \cline{2-5}
 &
  S &
  U &
  \textit{HM} &
  \textit{ZSL} \\ \hline
 Base$^{*}$ & 49.14 & 17.46 & 25.77 & 19.03 \\
 Base$^{*}$ + QDMA & 44.32 & 18.86 & 26.46 & 28.78 \\
 Base$^{*}$ + QDMA + CSGM (\(\boldsymbol{\mathcal{V}_{c}}\)$^{\#}$) & 46.34 & 23.64 & 31.31 & 28.44 \\
\hline
\rowcolor{lightgray!30}
 DAAN & 52.38 & 23.48 & \textbf{32.42} & \textbf{31.09} \\
\hline \hline
\multicolumn{4}{l}{$^{*}$baseline model comprising with MLP and cross-attention.}\\
\multicolumn{4}{l}{$^{\#}$convergence rate used only.}
\end{tabular}
\end{threeparttable}
}
\end{table}

\begin{figure}[t]  
 \centering      
  \subfloat[$QDMA$]
 {
   \label{fig:subfig1}\includegraphics[width=0.459\linewidth]{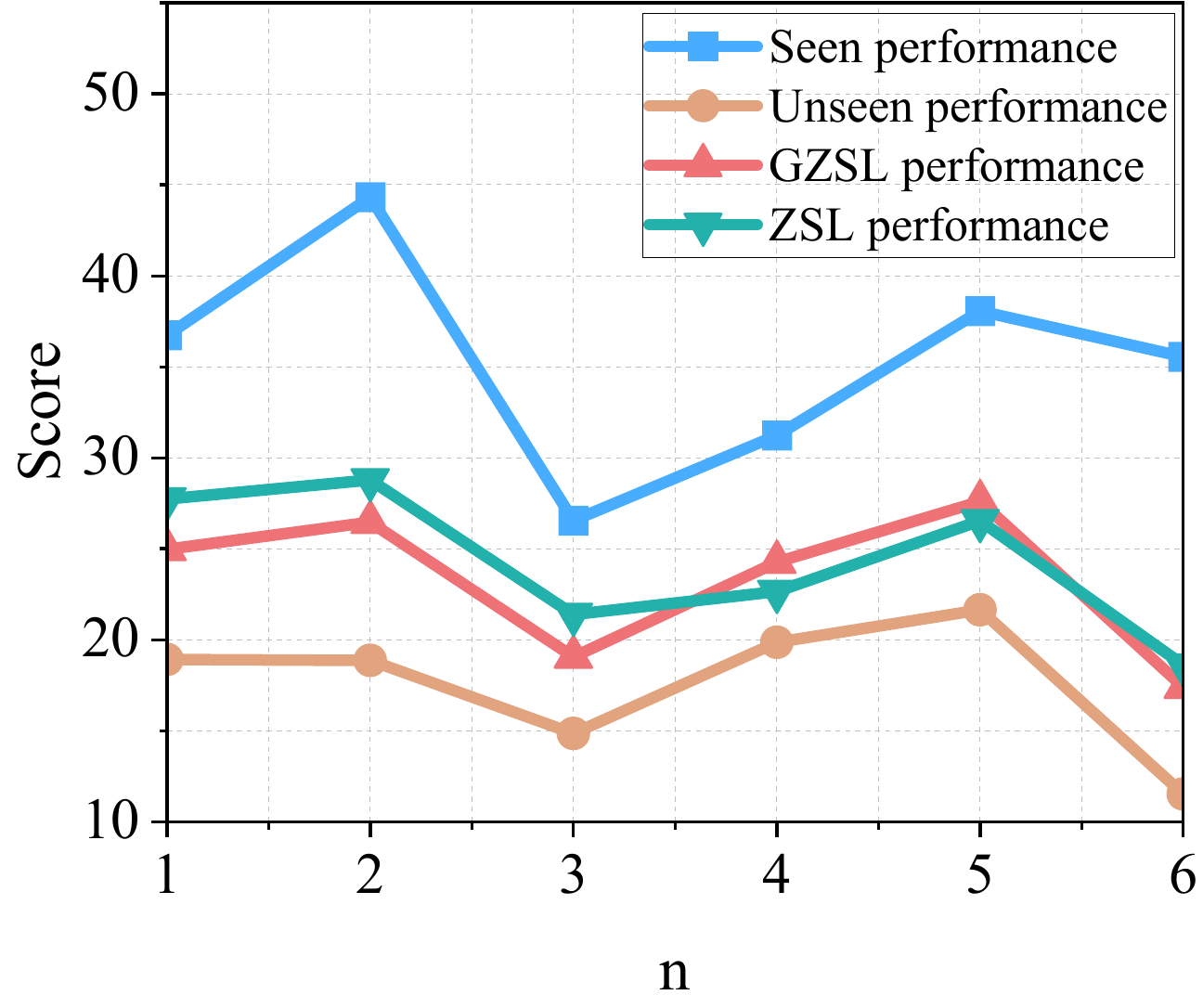}
 }
  \subfloat[$CSGM$]
 {
   \label{fig:subfig2}\includegraphics[width=0.451\linewidth]{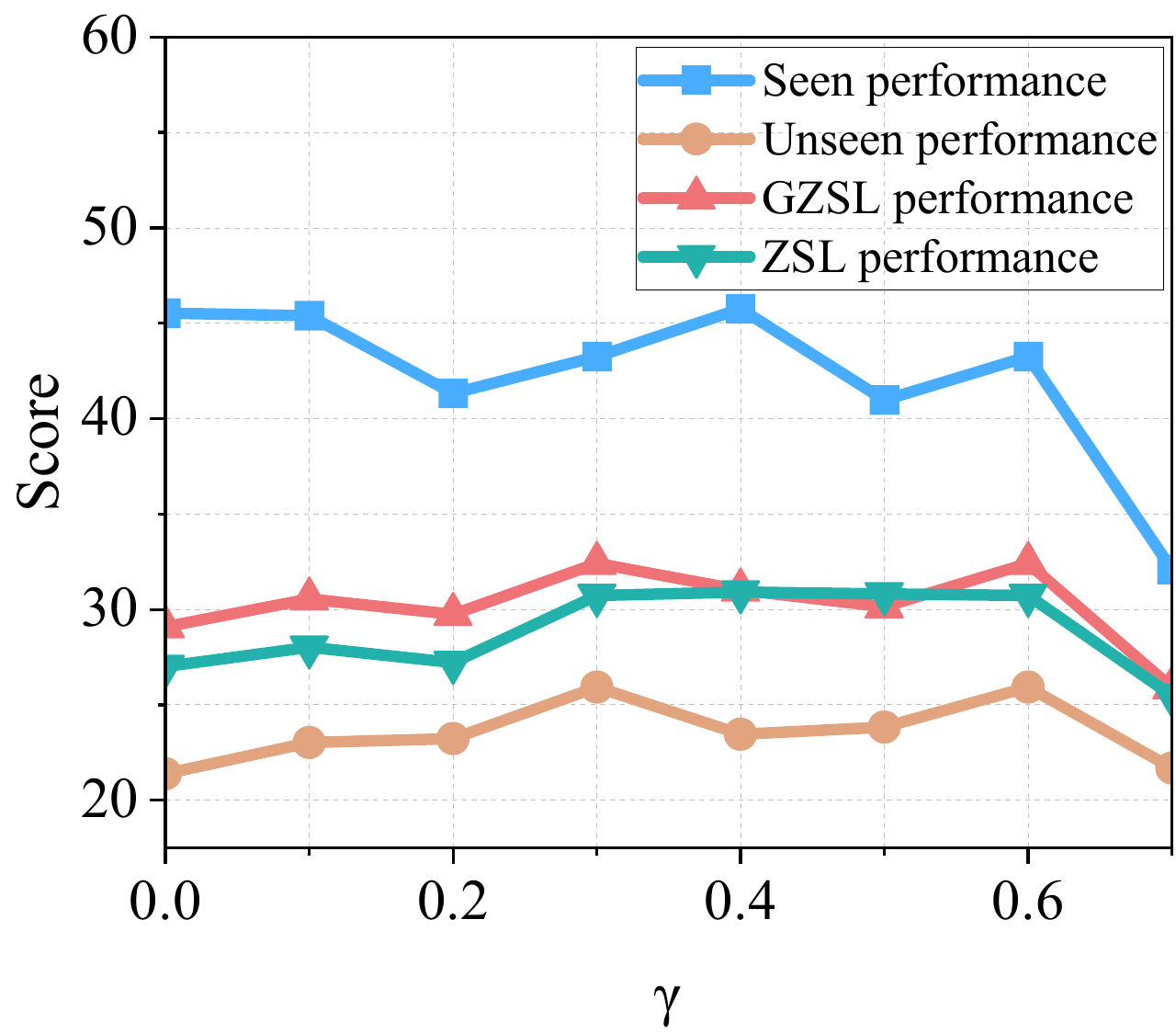}
 }
 \caption{Component Performance}
 \label{fig3}      
\end{figure}

% Of course, arriving the best performance compared to baseline acquires suitable parameters setting, Fig. (\ref{fig3}) shows the results of varying the hyperparameter \(\boldsymbol{n}\) of the QDMA module under the group “Base + QM” in Table \ref{tab:ablation_ucf-gzsl} while fixing the \(\boldsymbol{k}\) to 3 based on UCF dataset. We found that the value of \(\boldsymbol{n}\) should not be set too high. Typically, an appropriate value for n can be found within the range of $2$ to $5$. In Fig. (\ref{fig4}), we add the CSGM block to the model shows the best performance in Fig. (\ref{fig3}). We fix \(\boldsymbol{\mu}\) to $0.6$ and change the value of \(\boldsymbol{\gamma}\). The role of the hyperparameter \(\boldsymbol{\gamma}\) is to prevent the contribution rates calculated from differing too greatly, which could otherwise have a significant impact on the gradient. our results shows that \(\boldsymbol{\gamma}\) is essential for Gradient Modulation and the \(\boldsymbol{\gamma}\) between $0.2$ and $0.6$ brings significant benefit to (G)ZSL and ZSL performance. Additionally, this further demonstrates that within an appropriate parameter range, the CSGM module plays an indispensable role in both (G)ZSL and ZSL.

To further analyze the performance trends of QDMA and CSGM under varying parameters, we conducted experiments by adjusting $\boldsymbol{n}$ and $\boldsymbol{\gamma}$ which correspond to the QDMA and CSGM components, respectively. The experimental results are presented in Fig. (\ref{fig3}). As shown in Fig. (\ref{fig3}), the ZSL and GZSL scores of DAAN fluctuate between $25$ and $30$ under different parameters, demonstrating the robustness of our model's performance. Specifically, in Fig. (\ref{fig:subfig1}), the (G)ZSL scores of DAAN reach their maxima of $27.59$ and $26.54$ respectively when \(\boldsymbol{n} = 5\). This indicates that an overly small or overly large number of the Dilated convolutional layers in the QDMA module will lead to the loss and redundancy of modal temporal information, resulting in poor performance. In Fig. (\ref{fig:subfig2}), the scores of all indicators of DAAN first increase and then decrease as the lower bound of the contribution rate $\boldsymbol{\gamma}$ in CSGM increases, peaking at \(\boldsymbol{\gamma} = 0.3\). This demonstrates that an excessively small lower bound may cause excessive modulation of the modal gradient, thereby overly diminishing the contribution of the optimal modality. Conversely, an excessively high lower bound may counteract the intended effect of gradient modulation, ultimately leading to modality imbalance.  

\subsection{\textbf{Limitations \& Future work}} In this experiment, we only utilized three widely used datasets for model training and evaluation, and the performance of the model on other forms of datasets remains unknown. Additionally, video samples often suffer from modality loss, while DAAN is currently applicable only to datasets with complete audio-visual modalities. For future work, our architecture could be extended to video classification tasks with significant modality diversities and relatively few samples. Moreover, it could be integrated with feature extraction from pre-trained models to achieve improved performance.

\section{Conclusion}
\label{Conclusion}
In conclusion, we propose a Discrepancy-Aware Attention Network (DAAN) to enhance Audio-Visual Zero-Shot Learning (ZSL). In this work, we integrate a QDMA unit to minimize redundant information in the high-quality modality and a CSGM block to dynamically adjust gradient magnitudes, addressing content discrepancies across samples. To further refine the learning process, modality contributions are evaluated by incorporating both optimization strategies and convergence rates, ensuring more precise gradient modulation within the CSGM block. Experimental results show that DAAN achieves state-of-the-art performance on the VGGSound, UCF101, and ActivityNet benchmark datasets. Additionally, ablation studies validate the individual effectiveness of each proposed component in the architecture.

\bibliographystyle{IEEEbib} % 这一行代码指定了文献引用的格式样式。需要按照期刊、会议对应的格式要求来填写
% \bibliography{reference} % 这一行指定了参考文献数据库的文件名，在这个例子中，文件名是reference.bib，所以需要填写为“文件名是reference”

\end{document}